\pdfoutput=1

\documentclass[11pt]{article}

\usepackage[]{acl}

\usepackage{times}
\usepackage{latexsym}

\usepackage[T1]{fontenc}

\usepackage[utf8]{inputenc}

\usepackage{microtype}
\usepackage{booktabs}
\usepackage{graphicx}
\usepackage{multirow}
\usepackage{amssymb}

\renewcommand{\thepage}{\arabic{page}}
\usepackage{fancyheadings}
\fancypagestyle{specialfooter}{%
  \fancyhf{}
  
  \fancyfoot[C]{\normalsize \thepage\\ \vspace{3pt}
\small\textit{The 22nd Workshop on Biomedical Natural Language Processing and BioNLP Shared Tasks}, pages 468–477\\
July 13, 2023 ©2023 Association for Computational Linguistics}
}
\title{Overview of the BioLaySumm 2023 Shared Task on Lay Summarization of Biomedical Research Articles}

\author{Tomas Goldsack$^{1}$, Zheheng Luo$^{2}$, Qianqian Xie$^{2}$, \\ \textbf{Carolina Scarton}$^{1}$, \textbf{Matthew Shardlow}$^{3}$, \textbf{Sophia Ananiadou}$^{2}$, \textbf{Chenghua Lin}$^{1}$ \\
        $^{1}$University of Sheffield, $^{2}$University of Manchester,
        $^{3}$Manchester Metropolitan University \\
        \texttt{\{tgoldsack1, c.lin, c.scarton\}@sheffield.ac.uk}\\
        \texttt{\{zheheng.luo, qianqian.xie, sophia.ananiadou\}@manchester.ac.uk}\\
        \texttt{m.shardlow@mmu.ac.uk}}

\begin{document}
\thispagestyle{specialfooter}

\maketitle
\begin{abstract}

This paper presents the results of the shared task on Lay Summarisation of Biomedical Research Articles (BioLaySumm), hosted at the BioNLP Workshop at ACL 2023.  
The goal of this shared task is to develop abstractive summarisation models capable of generating ``lay summaries'' (i.e., summaries that are comprehensible to non-technical audiences) in both a controllable and non-controllable setting.
There are two subtasks: 1) Lay Summarisation, where the goal is for participants to build models for lay summary generation only, given the full article text and the corresponding abstract as input; and
2) Readability-controlled Summarisation, where the goal is for participants to train models to generate both the technical abstract and the lay summary, given an article’s main text as input.
In addition to overall results, we report on the setup and insights from the BioLaySumm shared task, which attracted a total of 20 participating teams across both subtasks.

\end{abstract}

\section{Introduction}
Biomedical publications report upon the latest research concerning prominent health-related topics, ranging from common illnesses to global pandemics~\cite{wang-etal-2020-cord}. Accordingly, the content of these publications is of interest to a wide variety of audiences, including researchers, medical professionals, journalists, and even members of the public. However, the highly technical and specialist language used within such articles typically makes it difficult for non-expert audiences to understand their contents. This results in useful knowledge and findings having limited accessibility to the general public~\cite{guo2021automated,goldsack-etal-2022-making,luo-etal-2022-readability}.

Abstractive summarisation models can be used to generate a concise summary of an article, capturing its most salient points using words and sentences that do not necessarily appear in the original text of the article.
As such, these models have the potential to make highly technical documents accessible to a much wider audience through the generation of  ``lay summaries''  --- more readable summaries consisting largely of background information and containing minimal technical terminology~\cite{guo2021automated,goldsack-etal-2022-making,luo-etal-2022-readability}.

The BioLaySumm shared task\footnote{\url{https://biolaysumm.org}} focuses on the abstractive summarisation of biomedical articles whilst placing an emphasis on controllability and ensuring comprehensibility for non-expert audiences. Through this shared task, we aim to foster increased research interest in Lay Summarisation (in both controllable and non-controllable settings), enabling further progression for novel model development and high-quality dataset construction. In turn, we hope this will help to broaden the accessibility of technical texts to non-specialist audiences and to drive progress towards more usable and effective abstractive summarisation models for the biomedical domain with the ability to cater to audiences possessing different levels of expertise.

In this paper, we present the results of the first BioLaySumm shared task, hosted by the BioNLP Workshop at ACL 2023. We cover the task formulation (\S\ref{sec:tasks}), datasets (\S\ref{sec:data}), and evaluation procedure (\S\ref{sec:eval}), before providing a description of the participating systems, overall results, and notable insights (\S\ref{sec:submission}).

\section{Task Description} \label{sec:tasks}
The shared task is composed of two separate subtasks, focusing on 1) the generation of summaries more suitable for a lay audience (Lay Summarisation), and 2) the development of controllable summarisation models capable of catering to audiences with different levels of expertise (Readability-controlled Summarisation).

\subsection{Subtask 1: Lay Summarisation}


Given an article’s abstract and main text as input, the goal is for participants to train a model (or models) to generate the lay summary. Two separate datasets, \textbf{PLOS} and \textbf{eLife} (derived from the eponymous biomedical journals), were provided for model training and will be used for evaluation (more details on datasets are given in \S\ref{sec:data}). For the evaluation, we average submission performance across both datasets.

For this task, we allowed submissions to be generated from either two separate summarisation models (i.e., one trained on each dataset) or a single unified model (i.e., trained on both datasets). Participants were required to indicate which approach was taken for each submission, in addition to whether or not they made use of additional training data (i.e., data not provided specifically for the task).


\subsection{Subtask 2: Readability-controlled Summarisation}
Given the main text of an article as input, the goal is for participants to train a model (or models) to generate both the technical abstract and the lay summary. A single dataset, \textbf{PLOS}, is provided for training and evaluation. 
We allowed submissions to use multiple ensemble models but still generate technical summary and the lay summary from the same model, and also one single main model with different output layers to generate two different summary types.
As with subtask 1, participants are required to indicate whether or not they made use of additional training data for each submission.
For the evaluation, we average submission performance across both summary types.

\section{Datasets} \label{sec:data}

The datasets used within each subtask are based on the previous works of \citet{goldsack-etal-2022-making} and \citet{luo-etal-2022-readability}, and are derived from two different biomedical publications: \textbf{Public Library of Science (PLOS)} and \textbf{eLife}. Each dataset consists of research articles, their technical abstracts, and their expert-written lay summaries.
As detailed in \S\ref{sec:tasks}, each form of summary within these datasets (i.e., abstract and lay summary) has a different utility in each subtask. The lay summaries of each dataset also exhibit numerous notable differences in their characteristics, with eLife's lay summaries being longer, more abstractive, and more readable than those of PLOS.
Furthermore, for PLOS, lay summaries are author-written, and articles are derived from 5 peer-reviewed journals covering Biology, Computational Biology, Genetics, Pathogens, and Neglected Tropical Diseases. For eLife, lay summaries are written by expert editors (in correspondence with authors), and articles are derived from the peer-reviewed eLife journal that covers all areas of the life sciences and medicine. For more detailed analysis of dataset content, please refer to \citet{goldsack-etal-2022-making}.

\begin{table}[]
    \centering
    \begin{tabular}{lcccc}
         \hline
         \textbf{Dataset} & \textbf{Subtask} & \textbf{\# Train} & \textbf{\# Val} & \textbf{\# Test}   \\ \hline
         eLife & 1 & 4,346 & 241 & 142 \\
         PLOS & 1, 2 & 24,773 & 1,376 & 142* \\ \hline
    \end{tabular}
    \caption{Data split sizes for each dataset. * denotes that this split is different for each subtask.}
    \label{tab:data}
\end{table}

Table \ref{tab:data} summarises the data split information for both datasets. Note that the training and validation sets used for both datasets are equal to those published in \citet{goldsack-etal-2022-making}. Furthermore, that the training and validation splits of PLOS are the same for both subtasks.


Alternatively, we collect new test splits for both PLOS and eLife data using more recently published articles from each respective journal.
The test data for Subtask 1 is composed of 142 PLOS articles and 142 eLife articles. The test data for Subtask 2 is composed of 142 PLOS articles (however, these are different from those used in Subtask 1).

In utilising these datasets for our task, we hope to enable the training of abstractive summarisation models that are capable of generating lay summaries for unseen articles covering a wide range of biomedical topics, enabling the significance of new, important publications to be effectively communicated to non-expert audiences.


\section{Evaluation}\label{sec:eval}



For both subtasks, we evaluate summary quality according to three criteria - \textit{Relevance}, \textit{Readability}, and \textit{Factuality} - where each criterion is composed of one or more automatic metrics:

\begin{itemize}
  \item \textit{Relevance}: ROUGE-1, 2, and L \citep{lin-2004-rouge} and BERTScore \citep{ZhangKWWA20}.
  \item \textit{Readability}: Flesch-Kincaid Grade Level (FKGL) and Dale-Chall Readability Score (DCRS).
  \item \textit{Factuality}: BARTScore \citep{NEURIPS2021_e4d2b6e6}, fine-tuned on our respective datasets (as has proven effective in recent work \citep{koh-etal-2022-far}).\footnote{A fine-tuned version of the FactCC \citep{kryscinski-etal-2020-evaluating} metric was also originally included for Factuality evaluation. However, preliminary testing found that it did not provide a reliable indication of factual correctness for the task.}
\end{itemize}

For Subtask 1, the scores calculated for each metric are the average of those calculated independently for the generated lay summaries of PLOS and eLife. The aim is to maximise the scores for Relevance and Factuality metrics and minimise scores for Readability metrics.

For Subtask 2, the scores presented for each metric are the average of those calculated independently for the generated abstracts and lay summaries. Notably, for Readability metrics in this subtask, we calculate the \textit{absolute difference} between the scores of generated summary and target summary pairs (rather than simply using the scores obtained for generated summaries, as in subtask 1). The aim is to maximise the scores for Relevance and Factuality metrics and minimise the absolute difference scores calculated for Readability metrics.

Following the submission deadline for each subtask, an overall ranking is calculated based on the cumulative rank of the evaluation criteria, where a lower overall ranking equates to better overall performance). To produce a criterion ranking, we apply min-max normalisation to the scores of each metric, before averaging across metrics within each evaluation criterion.

\section{Shared Task Submissions} \label{sec:submission}

For both subtasks, we include a baseline system based on BART-base \citep{lewis-etal-2020-bart} in order to provide a simple, widely-used benchmark with which submission performance can be compared. For subtask 1, this baseline system is composed of two separate BART models, trained independently on the PLOS and eLife datasets. For subtask 2, the baseline system is a controllable BART model, trained to generate either the abstract or lay summary of an article based on the inclusion of control tokens prepended to the input document ([\textsc{Abstract}] and [\textsc{Summary}], respectively). Participating teams for each subtasks were allowed to make a maximum of 3 submissions in total.

\subsection{Submissions to Subtask 1}

\subsubsection{Systems Overview}
\begin{table*}[ht]
    \centering
    \resizebox{1.0\textwidth}{!}{
    \begin{tabular}{lcccccccccccc}
        \hline \multirow{2}{*}{\textbf{Team}}  & \multirow{2}{*}{\textbf{\#}} & \multirow{2}{*}{\textbf{+}} & \multicolumn{4}{c}{\textbf{Relevance}} && \multicolumn{2}{c}{\textbf{Readability}} &&  \textbf{Factuality}  \\ \cline{4-7} \cline{9-10} \cline{12-12}
        &&& \textbf{R-1} & \textbf{R-2} & \textbf{R-L} & \textbf{BERTs} && \textbf{FKGL} & \textbf{DCRS} && \textbf{BARTs} \\
           \hline
           \textbf{himil} & 2 & $\times$  & \textbf{49.46} & 15.68 & 45.91 & 85.85 && 13.17 & 10.14 && -2.41 \\
           \textbf{Path. Dynamics} & 2 & $\times$ & 49.38 & 15.93 & 45.97 & 85.93 && 13.10 & 10.12 && -2.33 \\
           \textbf{Marsfield\_SDS} & 2 & \checkmark & 49.33 & 16.24 & \textbf{46.15} &  86.10 && 12.55 & 9.84 && -2.25 \\
           \textbf{LHS712EE} & 2 & $\times$ & 49.27 & 15.75 & 45.84 & 86.57 && 13.31 & 10.22 && -1.12 \\
           \textbf{APTSumm} & 2 & \checkmark & 48.32 & 14.91 & 45.41 & 84.30 && 12.22 & 9.00 && -3.41 \\
           \textbf{VBD-NLP} & 2 & $\times$ & 48.29 & 14.69 & 45.02 & 85.71 && 12.29 & 10.09 && -1.74 \\
           \textbf{MDC} & 1 & $\times$ & 48.22 & 15.53 & 44.85 & \textbf{87.07} && 12.94 & 10.21 && -1.18 \\
           \textbf{HanyangLab} & 2 & $\times$ & 48.18 & 14.18 & 44.20 & 85.83 && 12.83 & 10.50 && -1.92 \\
           \textbf{Arizona Sky} & 2 & $\times$ & 48.11 & 14.50 & 44.42 & 85.75 && 12.36 & 10.12 && -1.80 \\
           \textbf{ViNLPSum} & - & - & 47.97 & 15.46 & 44.85 & 85.77 && 12.77 & 9.68 && -2.01 \\
           \textbf{GRASUM} & 1 & \checkmark & 47.69 & \textbf{16.76} & 44.30 & 86.01 && 12.73 & 10.50 && -2.33 \\
           \textbf{IKM\_Lab} & 1 & $\times$ & 47.44 & 14.45 & 44.31 & 85.66 && 11.84 & 9.84 && -2.34 \\
           \underline{\textbf{baseline}} & 2 & $\times$ & 46.96 & 14.45 & 43.71 & 86.42 && 12.07 & 10.25 && \textbf{-0.83} \\
           \textbf{NCUEE-NLP} & 2 & $\times$ & 45.87 & 13.41 & 41.84 & 85.47 && 12.94 & 10.48 && -2.71 \\
           \textbf{HUST-NLP} & - & - & 43.29 & 12.25 & 39.22 & 85.30 && 12.69 & 10.79 && -1.91 \\
           \textbf{IITR} & 1 & $\times$ & 42.81 & 11.17 & 32.15 & 85.14 && \textbf{10.70} & \textbf{8.45} && -1.79 \\
           \textbf{noobitA} & - & - & 42.75 & 11.50 & 39.45 & 85.77 && 12.72 & 10.41 && -1.83 \\
           \textbf{LaSTUS-FBK} & 1 & $\times$ & 39.98 & 10.46 & 36.53 & 84.79 && 15.63 & 11.25 && -2.33 \\
           \textbf{ISIKSumm} & 1 & $\times$ & 37.06 & 9.04 & 34.48 & 82.55 && 12.13 & 10.02 && -3.92 \\
           \textbf{BITSpi} & 2 & $\times$ & 37.03 & 10.77 & 33.66 & 84.97 && 11.77 & 10.97 && -2.97 \\
           \textbf{nippon} & - & - & 36.53 & 8.82 & 33.62 & 83.44 && 12.80 & 10.42 && -2.28
    \end{tabular}
    }
    \caption{Subtask 1 leaderboard - all metrics. The \textbf{\#} column denotes the number of models used - 1 (unified) or 2 (one for each dataset), and the \textbf{+} column denotes the use of additional training data. "-" indicates that the corresponding information was not provided. \textbf{R} = ROUGE F1, \textbf{BERTs} = BERTScore, \textbf{FKGL} = Flesch-Kincaid Grade Level, \textbf{DCRS} = Dale-Chall Readability Score, \textbf{BARTs} = BARTScore.
    }
    \label{tab:st1_results}
\end{table*}

\begin{table*}[ht]
    \centering
    \resizebox{0.8\textwidth}{!}{
    \begin{tabular}{clcccc}
        \hline \textbf{Pos.} & \textbf{Team} & \textbf{Relevance} & \textbf{Readability} & \textbf{Factuality} & \textbf{Sum} \\
        \hline
        1 & \textbf{MDC} & 3 & 10 & 3 & \textbf{16} \\
        2 & \underline{\textbf{baseline}} & 9 & 8 & \textbf{1} & 18\\ 
        3* & \textbf{Marsfield\_SDS} & 2 & 6 & 11 & 19 \\
        3* & \textbf{VBD-NLP} & 8 & 7 & 4 & 19 \\
        5 & \textbf{LHS712EE} & \textbf{1} & 18 & 2 & 21 \\
        6 & \textbf{ViNLPSum} & 7 & 5 & 10 & 22 \\
        7 & \textbf{IITR} & 17 & \textbf{1} & 5 & 23 \\
        8 & \textbf{Arizona Sky} & 10 & 9 & 6 & 25 \\
        9 & \textbf{Path. Dynamics} & 4 & 11 & 14 & 29 \\
        10 & \textbf{IKM\_Lab} & 11 & 3 & 16 & 30 \\
        11 & \textbf{himil} & 5 & 13 & 17 & 34 \\
        12* & \textbf{APTSumm} & 13 & 2 & 20 & 35 \\
        12* & \textbf{GRASUM} & 6 & 16 & 13 & 35 \\
        14 & \textbf{noobitA} & 16 & 13 & 7 & 36 \\
        15 & \textbf{HanyangLab} & 12 & 17 & 9 & 38  \\
        16 & \textbf{HUST-NLP} & 15 & 20 & 8 &  43\\
        17 & \textbf{ISIKSumm} & 21 & 4 & 21 & 46 \\
        18 & \textbf{nippon} & 20 & 15 & 12 & 47 \\
        19 & \textbf{NCUEE-NLP} & 14 & 29 & 18 & 51 \\
        20 & \textbf{BITSpi} & 19 & 14 & 19 & 52 \\
        21 & \textbf{LaSTUS-FBK} & 19 & 21 & 15 & 54\\
    \end{tabular}
    }
    \caption{Subtask 1 leaderboard - criteria rankings.}
    \label{tab:st1_results-rank}
\end{table*} 
    

Subtask 1 attracted a total of 20 participating teams, between them making a total of 49 submissions. A brief explanation of the modelling approach taken by each team is given below:\footnote{Note that we were unable to get a response from every team describing their modelling approach, hence there are some teams missing from this section.} 

\paragraph{LHS712EE \normalfont\citep{LHS712EE}}  The team employed a BART model for eLife and a Longformer Encoder-Decoder (LED) model \citep{Beltagy2020Longformer} for PLOS, whilst also experimenting with optimising memory usage.

\paragraph{GRASUM \normalfont\citep{grasum}} Standing for \textbf{G}rounded, \textbf{R}eferenced, and \textbf{A}nnotated \textbf{SUM}maries, this team's method combines approaches from retrieval augmentation, offline RL, and controlled generation, using a LED model  with 16k input limit as the base model. A ``grounding'' step enhances each document with content retrieved from scientific abstracts, Wikipedia, simple Wikipedia, and UMLS that is appended to input, in addition the bibliographic reference string of the source document (obtained from CrossRef). An additional ``annotation" step annotates each source document with control tokens that indicate whether the corresponding summary achieves higher or lower than the median score for each of the task evaluation aspects (given in \S\ref{sec:eval}).

\paragraph{BITSpi} This team's method involves fine-tuning two separate BART models on pre-processed versions of each dataset. Specifically, stopwords are removed from the input data, and abbreviations are substituted for their full forms using a medical dictionary.

\paragraph{APTSum \normalfont\citep{APTSumm}} A three-step approach is adopted by this team, leveraging the SimCLS contrasting learning framework \citep{liu-liu-2021-simcls}. Specifically, they first perform content selection, identifying the Abstract and Introduction as best model input, before generating candidate summaries using BART, followed section-wise re-ranking using a RoBERTa-base model to capture section-based salience information.

\paragraph{LaSTUS-FBK} This team used a multi-stage unified approach, first cleaning the data via reference removal and acronym resolution. Extractive summarisation based on similarity-based sentence classification is then used to shorten the input before the resulting text is enhanced with the injection of complex concept definitions from Wikipedia. Finally, abstractive summarisation is performed using a fine-tuned BART model pre-trained on PubMed on a dataset-balanced sample of the training data (4K training instances from each dataset).    

\paragraph{Marsfield\_SDS \normalfont\citep{CSIRO-Data61}} Using two fine-tuned FLAN-T5 models (one for each dataset) as the backbone of their experiments, this team experimented with different data augmentation strategies including the use of ChatGPT for paraphrasing existing lay summaries.



\paragraph{VBD-NLP \normalfont\citep{VBD-NLP}} This team's method is based on the combined use of sequence-to-sequence model BioBART \citep{yuan-etal-2022-biobart} and FACTORSUM \citep{fonseca-etal-2022-factorizing}, a factorized energy-based model that aims to identify the most important input content, enabling more effective processing of long documents. Additional experimentation with handling length as well as utilising other Pretrained Language Models (PLMs) was also carried out.

\paragraph{MDC \normalfont\citep{MDC}} This team focused on comparing the performance of general-purpose GPT models (e.g., ChatGPT) with in-domain GPT models (e.g., BioGPT \citep{10.1093/bib/bbac409}). Additionally, they experimented with zero-shot and few-shot prompting, as well as fine-tuning different models. 

\paragraph{Pathology Dynamics \normalfont\citep{PathologyDynamics}} The team experimented with multiple different approaches based on BART and T5 models including methods of content selection, the use of efficient attention mechanisms (to better process long documents), and the zero-shot simplification of model outputs. Of those tested, the approach that achieved the best overall performance was BART-large, pretrained on CNN-DM dataset, with inputs truncated to 1024 tokens.


\paragraph{IITR \normalfont\citep{IITR}} Also using BART and T5 models trained on both dataset, this team experiment with different methods of content selection and ordering.

\paragraph{Arizona Sky} This team first truncate input documents, before using them to train two separate BART base models. 


\paragraph{IKM\_Lab \normalfont\citep{IKM_Lab}} This team experimented with the use of a LED model trained on both datasets, as well as the adoption of different formats for including additional article information, such as keywords and section headings, in the input.  

\paragraph{NCUEE-NLP \normalfont\citep{NCUEE-NLP}} This team also made use of different models for each submission, including Primera \citep{xiao-etal-2022-primera}, a PEGASUS model \citep{pegasus2020} pretrained on PubMed, and a BART-large Longformer model.

\paragraph{himil} The team experimented with both BERT \citep{devlin-etal-2019-bert} and Longformer-based models, trained individually on each dataset. 



\subsubsection{Results}


Table \ref{tab:st1_results} presents the performance of the submission selected to appear on the leaderboard by each team according to the defined task metrics and Table \ref{tab:st1_results-rank} presents the rankings of these submissions (both overall and according to each individual criteria) following the application of the evaluation process described in \S\ref{sec:eval}.

In general, we find that more teams opted for the use of two models (10 out of 20), one for each of the two provided datasets, rather than a single unified model trained on both datasets (6 out of 20). Furthermore, the use of additional training data (i.e., data not provided as part of this task) to directly fine-tune models was relatively rare, with only 3 confirmed instances. However, all participants decided to make use of pre-trained language models (PLMs) in their submissions.
In terms of the specific models used, we find BART-based models (e.g., BART, Longformer Encoder-Decoder, etc.) to be a particularly popular choice amongst teams, being utilised by 11 out of 13 teams who provided detailed descriptions of their method.
Finally, we observe that several teams also chose to experiment with data preprocessing, implementing methods such as data cleaning, data annotation, and data augmentation with varying degrees of success. 

We find that the best overall system (i.e., that which achieved the lowest summed ranking across the three evaluation criteria) is that of team MDC, whose best submission utilises a single ChatGPT-based model (text-davinci-003) coupled with few-shot prompting to generate the lay summaries of both datasets, based on only the abstracts. Although this system does not achieve the best performance in any individual criteria, it achieves a strong performance for both Relevance and Factuality (ranking 3rd for both) whilst maintaining an above-average Readability ranking (10th). The fact that the cumulative rank of this system is equal to 16 is evidence that no model is able to achieve universally strong performance across all criteria (relative to other submissions). However, the fact that the top-ranking submission is based on only few-shot in-context learning (i.e., without any fine-tuning on the provided training data) suggests that Large Language Models have the potential to offer significant benefits for Lay Summarisation. 

Interestingly, this is the only submission to achieve a better cumulative rank than that of the BART baseline system (18th), which is shown to rank first for Factuality, and above average for the other two criteria (9th and 8th for Relevance and Readability, respectively). We originally suspected that a possible explanation for the baseline system's strong performance in terms of Factuality is a potential bias of BARTScore towards BART-based models. However, the leaderboard results do not seem to support this, with BART-based models being widely used and achieving a wide range of scores.

\begin{table*}[!ht]
    \centering
    \resizebox{1.0\textwidth}{!}{
    \begin{tabular}{lcccccccccccc}
        \hline \multirow{2}{*}{\textbf{Team}}  & \multirow{2}{*}{\textbf{\#}} & \multirow{2}{*}{\textbf{+}} & \multicolumn{4}{c}{\textbf{Relevance}} && \multicolumn{2}{c}{\textbf{Readability}} &&  \textbf{Factuality}  \\ \cline{4-7} \cline{9-10} \cline{12-12}
        &&& \textbf{R-1} & \textbf{R-2} & \textbf{R-L} & \textbf{BERTs} && \textbf{FKGL} & \textbf{DCRS} && \textbf{BARTs} \\
           \hline
        \textbf{LHS712EE} &2&$\times$& 44.17&	12.99&	40.53&	\textbf{85.49}&&	2.263	&0.9364&&	-1.1403 \\
           \textbf{NCUEE-NLP} &1&$\times$& \textbf{45.14}&	\textbf{14.02}&	\textbf{41.23}&	85.45	&&\textbf{2.047}	&0.9340&&	-2.1102 \\
           \textbf{Pathology Dynamics} &1&$\times$& 45.11&	13.82&	41.00&	85.32&&	2.106&	\textbf{0.8232}&&	-1.5682 \\
           \underline{\textbf{baseline}} &1&$\times$& 40.88&	11.63	&36.86	&\textbf{85.49}	&&2.396&	0.9312	&& \textbf{-0.9783}
    \end{tabular}
    }
    \caption{Subtask 2 leaderboard-all metrics. The \textbf{\#} column denotes the number of models used - 1 (unified) or 2 (one for each dataset), and the \textbf{+} column denotes the use of additional training data. \textbf{R} = ROUGE F1, \textbf{BERTs} = BERTScore, \textbf{FKGL} = Flesch-Kincaid Grade Level, \textbf{DCRS} = Dale-Chall Readability Score, \textbf{BARTs} = BARTScore.}
    \label{tab:control sum lb}
\end{table*}

\begin{table*}[!ht]
    \centering
    \resizebox{0.8\textwidth}{!}{
    \begin{tabular}{clcccc}
        \hline \textbf{Pos.} & \textbf{Team} & \textbf{Relevance} & \textbf{Readability} & \textbf{Factuality} & \textbf{Sum} \\
        \hline
        1* & \textbf{NCUEE-NLP} & 1 & 2 & 4 & 7 \\
        1*& \textbf{Pathology Dynamics} & 3 & 1 & 3 & 7 \\
        1*&\textbf{LHS712EE} &2& 3 & 2 & 7 \\
         4 & \underline{\textbf{baseline}} & 4 & 4 & 1 & 9\\
         
    \end{tabular}
    }
    \caption{Subtask 2 leaderboard - criteria rankings.}
    \label{tab:st2_results-rank}
\end{table*} 

Two teams tied for third in terms of overall ranking, with both Marsfield\_SDS and VBD-NLP achieving a cumulative rank of 19. Each of these teams adopted innovative and diverse strategies with their submissions. Marsfield\_SDS focused largely on data augmentation including the use of ChatGPT for generating lay summary paraphrases, resulting in particularly strong performance in terms of Relevance (2nd). Alternatively, VBD-NLP experimented with the use of the factorised energy-based model FACTORSUM, achieving a good all-rough performance across all criteria. Finally, the 5th placed submission of team LHS712EE is also worthy of note, obtaining the best rank for Relevance and 2nd best for Factuality.

\subsection{Submissions to Subtask 2}

\subsubsection{Systems Overview}
Three teams have made in total 7 attempts for Subtask 2. A brief description of their respecitve approaches are as following:

\paragraph{LHS712EE \normalfont\citep{LHS712EE}}
The team carried on with the LED \cite{Beltagy2020Longformer} model trained on the PLOS dataset from Subtask 1 to test the generalizability of their approach in generating lay summaries coupled with a pre-trained LED model for abstractive summaries. They later retrained the model using the abstract section of the dataset to improve performance in generating technical abstracts. 

\paragraph{Pathology Dynamics \normalfont\citep{PathologyDynamics}}
As the abstract with the most salient information is no longer
present in the input, to tackle the long context input, the team trained a base LSG model \cite{Condevaux2022LSGAE} and truncated each article to the first
4096 tokens for generating both abstracts and lay summaries. The model was then
trained on a merged dataset that uses each article twice, with one output having the lay summaries and the other having the abstract. They also reported using simplification procedures such as MUSS \cite{Martin2022MUSSMU} to enhance the lay summary or other instruction-following models such as T5 with different prefix for summarisation.

\paragraph{NCUEE-NLP \normalfont\citep{NCUEE-NLP}} 
This team made use of different models for each submission, including Primera, a PEGASUS model pre-trained on PubMed, and a BART-large Longformer model.

\subsubsection{Results}
In Table \ref{tab:control sum lb}, the performance of the submissions to Subtask 2 is shown on the leaderboard by each team according to the defined task metrics.  Table \ref{tab:st2_results-rank} presents  the overall and by individual metric rankings of these submissions following the application of the evaluation process described in \S\ref{sec:eval}. Due to the overall ranking scheme and the limited number of participants, we have all three teams ranked first while demonstrating advantages and disadvantages in different aspects.

All three teams utilise augmented transformers that can take longer input context, which significantly boosts the performance of Rouge score while also achieving smaller readability differences. We assume this is because longer input enables the models to see more lexicons that can be used to build summaries, resulting in a better chance to overlap with the reference summaries. However, these improvements do not necessarily promise higher results on LM-based metrics such as BERTScore and BARTScore on which the baseline method prevails.

It is worth noting that Team Pathology Dynamics used summaries generated from a LSG model simultaneously trained on both plain language as well as technical references and get output as a
hybrid of the lay summaries and abstracts. Their methods obtains the highest readability and joint overall highest scores, suggesting the limitation of the readability metrics used for evaluation. In addition, they reported that neither simplification model nor small-scale instruction-following models succeed to improve performance in this task.

In conclusion, none of the participating team secured a sweeping superiority across the three evaluated aspects, highlighting the challenge in readability-controlled summarisation on relatively small-scaled language models. Given that LLMs (Large Language Models) better align with human instructions \cite{Ouyang2022TrainingLM}, we expect future work to investigate their capabilities in the task.

\section{Conclusion}


The first BioLaySumm shared task was hosted at the BioNLP Workshop @ ACL2023 and consisted of two subtasks focusing on Lay Summarisation and Readability-controlled Summarisation, respectively. The task attracted a total of 20 teams, between them making 56 individual submissions across both subtasks. Submissions were evaluated according to three general criteria - Relevance, Readability, and Factuality --- with each criteria consisting of one or more automatic metrics. 

The results of both subtasks show that achieving strong performance for all three criteria (relative to other submissions) was particularly rare, attesting to the challenging nature of generating lay summaries for research articles in both controlled and non-controlled settings. Furthermore, when also taking into account the relatively strong performance of the BART baseline models (in particular for the Factuality component of our evaluation), this suggests that further research effort is required to develop truly usable models that can be reliably deployed in real-world settings. 

However, as demonstrated by highly-ranked teams MDS and Marsfield\_SDS (who obtain first and joint third-ranking submissions for subtask 1, respectively), recent developments in the abilities of both general-purpose and in-domain LLMs have the potential to offer significant benefits for the automatic generation lay summaries. As such, we expect that utilising such models for summary generation, data augmentation, and evaluation to be promising future directions for Lay Summarisation.

\bibliography{anthology,custom}
\bibliographystyle{acl_natbib}

\appendix
\section{Appendix}

\begin{table*}[ht]
    \centering
    \resizebox{0.85\textwidth}{!}{
    \begin{tabular}{lcccccccccc}
        \hline \multirow{2}{*}{\textbf{Team}} & \multicolumn{4}{c}{\textbf{Relevance}} && \multicolumn{2}{c}{\textbf{Readability}} &&  \textbf{Factuality}  \\ \cline{2-5} \cline{7-8} \cline{10-10}
        & \textbf{R-1} & \textbf{R-2} & \textbf{R-L} & \textbf{BERTs} && \textbf{FKGL} & \textbf{DCRS} && \textbf{BARTs} \\
           \hline
           \textbf{himil} & \textbf{1} & 0.864 & 0.983 & 0.730 && 0.491 & 0.601 && 0.488 \\
           \textbf{Path. Dynamics} & 0.994 & 0.895 & 0.987 & 0.748 && 0.476 & 0.596 && 0.515 \\
           \textbf{Marsfield\_SDS} & 0.990 & 0.935 & \textbf{1} & 0.785 && 0.364 & 0.496 && 0.541 \\
           \textbf{LHS712EE}      & 0.985 & 0.873 & 0.978 & 0.889 && 0.521 & 0.631 && 0.906 \\
           \textbf{APTSumm}       & 0.912 & 0.767 & 0.947 & 0.387 && 0.295 & 0.192 && 0.167 \\
           \textbf{VBD-NLP}         & 0.909 & 0.739 & 0.919 & 0.699 && 0.310 & 0.583 && 0.707 \\
           \textbf{MDC}           & 0.904 & 0.845 & 0.907 & \textbf{1} && 0.443 & 0.625 && 0.888 \\
           \textbf{HanyangLab}    & 0.901 & 0.675 & 0.861 & 0.726 && 0.422 & 0.730 && 0.649  \\
           \textbf{Arizona Sky}   & 0.896 & 0.715 & 0.876 & 0.708 && 0.324 & 0.592 && 0.685 \\
           \textbf{ViNLPSum}      & 0.885 & 0.826 & 0.907 & 0.712 && 0.408 & 0.438 && 0.618 \\
           \textbf{GRASUM}        & 0.863 & \textbf{1} & 0.868 & 0.765 && 0.401 & 0.732 && 0.516  \\
           \textbf{IKM\_Lab}      & 0.844 & 0.775 & 0.866 & 0.688 && 0.217 & 0.495 && 0.512 \\
           \underline{\textbf{BART Baseline}} & \underline{0.807} & \underline{0.709} & \underline{0.826} & \underline{0.856} && \underline{0.264} & \underline{0.641} && \underline{\textbf{1}} \\
           \textbf{NCUEE-NLP}    & 0.722 & 0.578 & 0.692 & 0.646 && 0.445 & 0.724 && 0.393 \\
           \textbf{HUST-NLP}     & 0.523 & 0.432 & 0.505 & 0.608 && 0.391 & 0.833 && 0.650 \\
           \textbf{IITR}         & 0.486 & 0.296 & 0     & 0.573 && \textbf{0} & \textbf{0} && 0.690 \\
           \textbf{noobitA}      & 0.481 & 0.338 & 0.521 & 0.712 && 0.399 & 0.698 && 0.678 \\
           \textbf{LaSTUS-FBK}   & 0.267 & 0.207 & 0.313 & 0.496 && 1     & 1 && 0.514 \\
           \textbf{ISIKSumm}     & 0.041 & 0.028 & 0.166 & 0     && 0.277 & 0.560 && 0 \\
           \textbf{BITSpi}       & 0.039 & 0.246 & 0.108 & 0.535 && 0.203 & 0.897 && 0.309 \\
           \textbf{nippon}       & 0     & 0     & 0.105 & 0.197 && 0.416 & 0.701 && 0.533
    \end{tabular}
    }
    \caption{Subtask 1 leaderboard - metric values normalised using min-max normalisation, so values range from 0-1.}
    \label{tab:st1_results_norm}
\end{table*}

\begin{table*}[ht]
    \centering
    \resizebox{0.85\textwidth}{!}{
    \begin{tabular}{lcccccccccc}
        \hline \multirow{2}{*}{\textbf{Team}} & \multicolumn{4}{c}{\textbf{Relevance}} && \multicolumn{2}{c}{\textbf{Readability}} &&  \textbf{Factuality}  \\ \cline{2-5} \cline{7-8} \cline{10-10}
        & \textbf{R-1} & \textbf{R-2} & \textbf{R-L} & \textbf{BERTs} && \textbf{FKGL} & \textbf{DCRS} && \textbf{BARTs} \\
           \hline
           \textbf{NCUEE-NLP}&	\textbf{1}	&\textbf{1}	&\textbf{1}	&0.764	&&\textbf{0}	&0.978	&&0\\
           \textbf{Pathology Dynamics}	&0.992&	0.916&	0.947&	0	&&0.169&	\textbf{0}	&&0.478 \\
           \textbf{LHS712EE}	&0.772	&0.569&0.839	&\textbf{1}&	&0.619	&1	&&0.856 \\
           \textbf{Baseline}&	0&	0	&0	&\textbf{1}&	&1	&0.954&&\textbf{1} \\
    \end{tabular}
    }
    \caption{Subtask 2 leaderboard - metric values normalised using min-max normalisation, so values range from 0-1.}
    \label{tab:st2_results_norm}
\end{table*}

\end{document}